\documentclass[10pt,twocolumn,letterpaper]{article}

\usepackage{cvpr}      % To produce the REVIEW version
\usepackage{booktabs}
\usepackage{amsfonts}
\usepackage{makecell}
\usepackage{amssymb}
\usepackage{amsmath}
\usepackage{multirow}
\usepackage{caption}
\usepackage{scalerel,graphicx,xparse}

\NewDocumentCommand\emojione{}{\scalerel*{\includegraphics{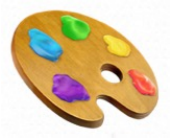}}{X}}

\definecolor{cvprblue}{rgb}{0.21,0.49,0.74}
\usepackage[pagebackref,breaklinks,colorlinks,allcolors=cvprblue]{hyperref}

\title{\emojione MVPainter: Accurate and Detailed 3D Texture Generation via Multi-View Diffusion with Geometric Control}
\author{
Mingqi Shao \and
Feng Xiong  \and
Zhaoxu Sun \and 
Mu Xu  \and
AMAP\\
{\tt\small \{mingqi.smq,szx430629,xumu.xm\}@alibaba-inc.com~}
}

\begin{document}
\maketitle

\begin{abstract}
Recently, significant advances have been made in 3D object generation. Building upon the generated geometry, current pipelines typically employ image diffusion models to generate multi-view RGB images, followed by UV texture reconstruction through texture baking. While 3D geometry generation has improved significantly, supported by multiple open-source frameworks, 3D texture generation remains underexplored. In this work, we systematically investigate 3D texture generation through the lens of three core dimensions: reference-texture alignment, geometry-texture consistency, and local texture quality. To tackle these issues, we propose \textbf{MVPainter}, which employs data filtering and augmentation strategies to enhance texture fidelity and detail, and introduces ControlNet-based geometric conditioning to improve texture-geometry alignment. Furthermore, we extract physically-based rendering (PBR) attributes from the generated views to produce PBR meshes suitable for real-world rendering applications. MVPainter achieves state-of-the-art results across all three dimensions, as demonstrated by human-aligned evaluations. To facilitate further research and reproducibility, we also release our full pipeline as an open-source system, including data construction, model architecture, and evaluation tools. Project page:
\href{https://amap-cvlab.github.io/MV-Painter/}{https://amap-cvlab.github.io/MV-Painter}

\end{abstract}    
\section{Introduction}
\label{sec:intro}

With the development of generative artificial intelligence, 3D content generation has gradually become a hot topic in computer vision and graphics. Among many tasks, generating a complete 3D object from a single image has attracted great attention due to its wide application prospects such as virtual reality\cite{jiang2024vr,li2024advances}, digital humans\cite{guo2020action2motion,huang2024humannorm,cao2024dreamavatar}, game design\cite{werning2024generative,bensadoun2024meta,hu2024game}, etc.. The current mainstream paradigm tends to decompose the 3D generation process into two subtasks\cite{bensadoun2024meta,zhang2024clay,zhao2025hunyuan3d}: first generate geometric shapes, and then perform texture synthesis. This "geometry-texture separation" strategy reduces the modeling complexity on the one hand, and enables researchers to optimize and study the two modules of geometry and texture separately on the other hand.

Geometry generation has made significant breakthroughs, and many methods have shown strong capabilities in single-image 3D reconstruction tasks\cite{li2024craftsman,li2025triposg,zhang2024g3pt,zhang20243d,wei2025octgpt}. Typical representatives such as TripoSG\cite{li2025triposg}, Hunyuan3D 2.0\cite{zhao2025hunyuan3d}, TRELLIS\cite{xiang2024structured} and Hi3dGen\cite{ye2025hi3dgen} have effectively improved the stability, structural complexity, and detail completeness of the generated shapes by directly modeling distribution in 3D space with diffusion or autoregressive models. These methods not only achieve high-quality geometric modeling but also have good generalization capabilities and can cope with multi-category and diverse object types. However, the maturity of geometry generation has also further exposed that texture generation is a shortcoming in the entire 3D-Gen process, and its effect often becomes a key factor in determining the upper limit of 3D generation quality.

\begin{figure}[tbp]
\centering 
\includegraphics[width=\linewidth]{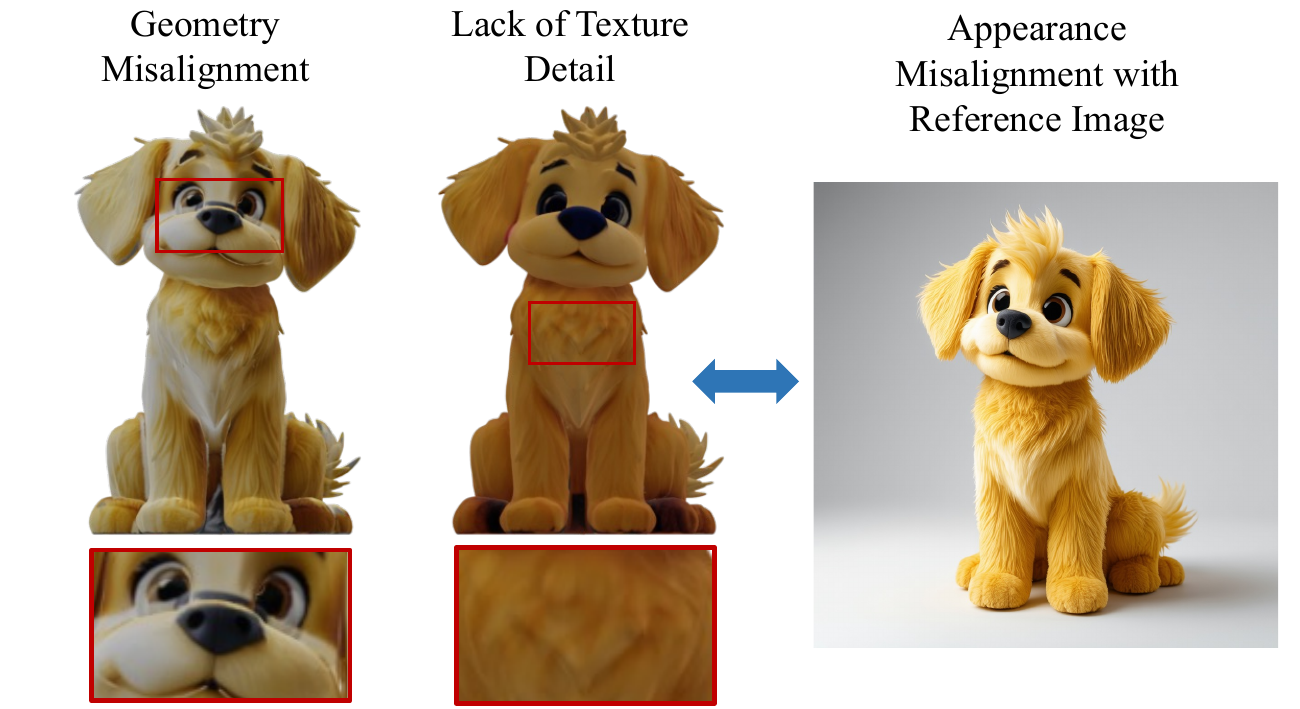} 
\caption{Three key challenges in existing 3D texture generation methods.} %最终文档中希望显示的图片标题
\label{fig:problem} %用于文内引用的标签
\end{figure}

At present, mainstream texture generation methods usually use 2D image diffusion models to generate RGB images\cite{bensadoun2024meta,huang2024mv,zhao2025hunyuan3d}, and then map them to 3D surfaces through a projection strategy. However, three core challenges remain unresolved in current approaches as shown in Fig. \ref{fig:problem}: \textbf{(1) Reference-texture alignment} — it is difficult to ensure that the generated texture accurately reflects the visual characteristics of the reference image, especially under varying lighting and occlusions;
\textbf{(2) Geometry-texture consistency} — aligning textures precisely with the 3D surface remains challenging;
\textbf{(3) Local texture quality }— many methods struggle to produce textures with sufficient details.

To address the aforementioned challenges, this work focuses on the three core issues of 3D texture generation and proposes a systematic modeling and evaluation framework. Our contributions are summarized as follows:

\begin{itemize}
    \item We construct a high-quality data processing pipeline to enhance the reference image alignment and improve the model's ability to learn fine-grained details.
    
    \item We introduce a geometry-guided architecture based on ControlNet, which incorporates explicit multi-modal signals such as normal and depth to improve the alignment between texture and geometry.
    
    \item We propose a VLM-based evaluation strategy aligned with human perception, covering three key dimensions: reference alignment, geometry-texture consistency, and local texture quality, and {MVPainter} achieves state-of-the-art performance across all three metrics.
\end{itemize}

In summary, this work addresses the three key challenges in 3D texture generation through a unified framework that integrates data preparation, model design, and evaluation. To encourage further research in this area, we release the complete MVPainter system, including data construction tools, training pipelines, and evaluation scripts.

\section{MVPainter Methodology}

\begin{figure*}[htbp]
\centering 
\includegraphics[width=\linewidth]{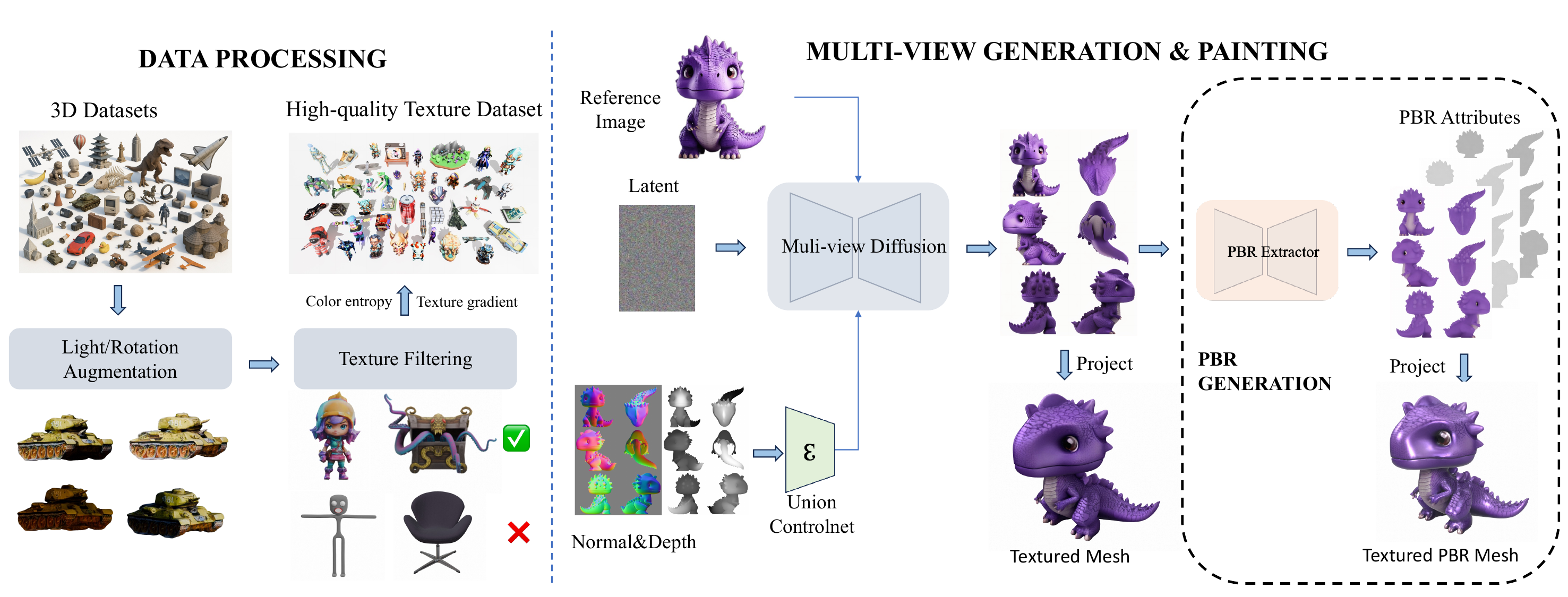} 
\caption{Overview of our framework. First, we apply data filtering and augmentation strategies to ensure that the training data contains sufficient detail and variations in lighting and viewpoint. Then, we leverage a ControlNet-based architecture to generate multi-view texture images that are structurally consistent with the 3D geometry. Finally, we introduce a dedicated PBR extraction module to estimate the basecolor, metallic, and roughness maps, which are projected back onto the 3D mesh to obtain a PBR-textured model.} %最终文档中希望显示的图片标题
\label{fig:overview} %用于文内引用的标签
\end{figure*}

In this report, we focus on the task of generating high-quality, geometry-consistent texture for a given 3D mesh and a single reference image. Formally, let the input be a reference image $\mathcal{I}_{\text{ref}}$ and the corresponding 3D geometry $\mathcal{G}$. Our objective is to build a texture generator $\mathcal{F}$ such that:
\begin{equation}
    \mathcal{F}(\mathcal{I}_{\text{ref}}, \mathcal{G}) \rightarrow \mathcal{M}_{\text{textured}}
\end{equation}
where $\mathcal{M}_{\text{textured}}$ denotes the final textured 3D mesh.

As mentioned in Sec.\ref{sec:intro}, this task poses several difficulties. To address these challenges, we propose targeted improvements from both data and network design perspectives. An overview of our approach is shown in Fig \ref{fig:overview}. First, we apply data filtering and augmentation strategies to ensure that the training data contains sufficient detail and variations in lighting and viewpoint. Then, we leverage a ControlNet-based architecture to generate multi-view texture images that are structurally consistent with the 3D geometry. Finally, we introduce a dedicated PBR extraction module to estimate the basecolor, metallic, and roughness maps, which are projected back onto the 3D mesh to obtain a PBR-textured model.

\subsection{Data Processing}
\label{sec:data_process}
In this section, we present a dedicated data processing and augmentation strategy tailored for 3D texture generation.  The proposed pipeline is designed to improve the model’s ability to align generated textures with the reference image while enhancing the quality of local texture details.

The input images used for 3D texture generation are typically captured from natural scenes and often exhibit strong lighting variations, including significant specular highlights and shadow boundaries. When such images are directly used as references, the multi-view diffusion model tends to overfit to lighting artifacts during training, resulting in unnatural or distorted textures. Although existing approaches such as Hunyuan3D 2.0\cite{zhao2025hunyuan3d} attempt to remove lighting effects through a delighting module, this process inevitably alters the original color distribution, leading to inconsistencies with the reference image. To address this issue, we propose a lighting augmentation-based data simulation strategy that constructs reference images under diverse lighting conditions. Specifically, we first render six target-view images under uniform lighting to eliminate specular and shadow artifacts. For the conditioning reference image, we employ a variety of lighting enhancements, including point lights, area lights, and HDR environment maps. For each object, we render 15 reference images, each illuminated by a randomly selected configuration from point lights, area lights, or a curated set of 100 HDRIs. This strategy enables the model to generate appearance-consistent and color-faithful textures, even when the reference input exhibits mismatched lighting distributions, thereby significantly improving its robustness to in-the-wild images. In addition, to address the issue that input images are often not captured from the canonical front-facing view, we introduce controlled perturbations in the rendering of reference images by varying the azimuth angle within the range of (–30°, 30°) and the elevation angle within (–10°, 30°).

Existing publicly available 3D datasets often contain a substantial number of texture samples with poor quality, such as those with monotonous colors, extremely low saturation, or missing surface patterns. Such low-quality data can hinder the model’s ability to learn high-fidelity texture representations, resulting in blurry outputs that lack structural sharpness and material expressiveness. To construct a high-quality training dataset, we filter the samples based on two key dimensions:
\begin{itemize}
    \item \textbf{Color Entropy}: Measures the diversity of color distribution within an image. A higher entropy value indicates richer and more varied color content.
    \item \textbf{Texture Complexity}: Quantifies the richness of local texture details using metrics such as image gradient distributions and frequency-domain energy.
\end{itemize}

\noindent\textbf{Color Entropy.} To quantitatively assess the diversity of color information in an image, we compute the \textit{color entropy} in the HSV color space. Specifically, given an input image, we first convert it from the RGB to the HSV color space to decouple chromatic and luminance information. For each of the three channels—hue (H), saturation (S), and value (V)—we compute a 256-bin normalized histogram and then calculate its Shannon entropy. The entropy for a channel \( c \in \{H, S, V\} \) is defined as:

\begin{equation}
\mathcal{H}(c) = - \sum_{i=1}^{256} p_i \log p_i
\end{equation}

where \( p_i \) denotes the normalized probability of the \( i \)-th bin in the histogram of channel \( c \). To obtain a single scalar representing the overall color richness of the image, we compute the average entropy across the three channels:

\begin{equation}
\mathcal{C}_{\text{color}} = \frac{1}{3} \left( \mathcal{H}(H) + \mathcal{H}(S) + \mathcal{H}(V) \right)
\end{equation}

This metric captures both chromatic and luminance variations in the image and serves as a robust indicator of color diversity. A higher color entropy value indicates that the image contains more varied and rich color distributions, which is desirable for training high-quality texture generation models.

\noindent\textbf{Texture Complexity.} To quantify the richness of local texture patterns, we measure the \textit{texture complexity} of an image using the average Sobel gradient magnitude over all pixels. Given an input image \( I \in \mathbb{R}^{H \times W \times 3} \), we first convert it to a grayscale image \( G \in \mathbb{R}^{H \times W} \). The horizontal and vertical gradients are then computed using the Sobel operator:

\begin{equation}
G_x = \text{Sobel}(G, \text{axis}=x), \quad
G_y = \text{Sobel}(G, \text{axis}=y)
\end{equation}
Then we combine the $G_X$ and $G_y$ into the total gradient magnitude $G_{mag}$:
\begin{equation}
G_{mag} = \sqrt{G_x(x, y)^2 + G_y(x, y)^2}
\end{equation}
The overall texture complexity score \( \mathcal{C}_{\text{texture}} \) is defined as the average gradient magnitude across all pixels:

\begin{equation}
\mathcal{C}_{\text{texture}} = \frac{1}{HW} \sum_{x=1}^{H} \sum_{y=1}^{W} G_{mag}
\end{equation}
$\mathcal{C}_{\text{texture}}$ captures the spatial frequency and structural variation in the image. Higher values of \( \mathcal{C}_{\text{texture}} \) indicate more complex and detailed textures, which are desirable for training models capable of generating rich surface appearances.

To assess the overall quality of each 3D object, we render its front and side views under uniform lighting conditions and compute the \textit{color entropy} and \textit{texture complexity} for each rendered image. We then compute a combined quality score for each object as:
\begin{equation}
\mathcal{C}_{\text{total}} = \lambda \cdot \mathcal{C}_{\text{color}} + \mathcal{C}_{\text{texture}}
\end{equation}
where  \( \lambda = 35 \) is a fixed weighting coefficient used to balance their scales.

All objects are ranked based on their total scores, and the top 100{,}000 objects are selected to construct the final high-quality texture dataset. This selection strategy ensures that only samples with sufficient color diversity and rich local texture patterns are used for model training, thereby improving the expressiveness and fidelity of the generated textures.

\subsection{Multi-view Generation with Geometric Control}
\label{sec:multi_view_generation}
MVPainter leverages a multi-view diffusion model to synthesize high-quality, geometry-consistent images from a single reference image and known 3D geometry. The overall architecture is illustrated on the right side of Fig.\ref{fig:overview}. Similar to methods such as Zero123++\cite{shi2023zero123++}, the six target views are arranged into a 3×2 image grid and treated as the generation target. The reference image and geometric priors serve as control signals to guide the multi-view diffusion generation.

\begin{figure}[bp]
\centering 
\includegraphics[width=1.1\linewidth]{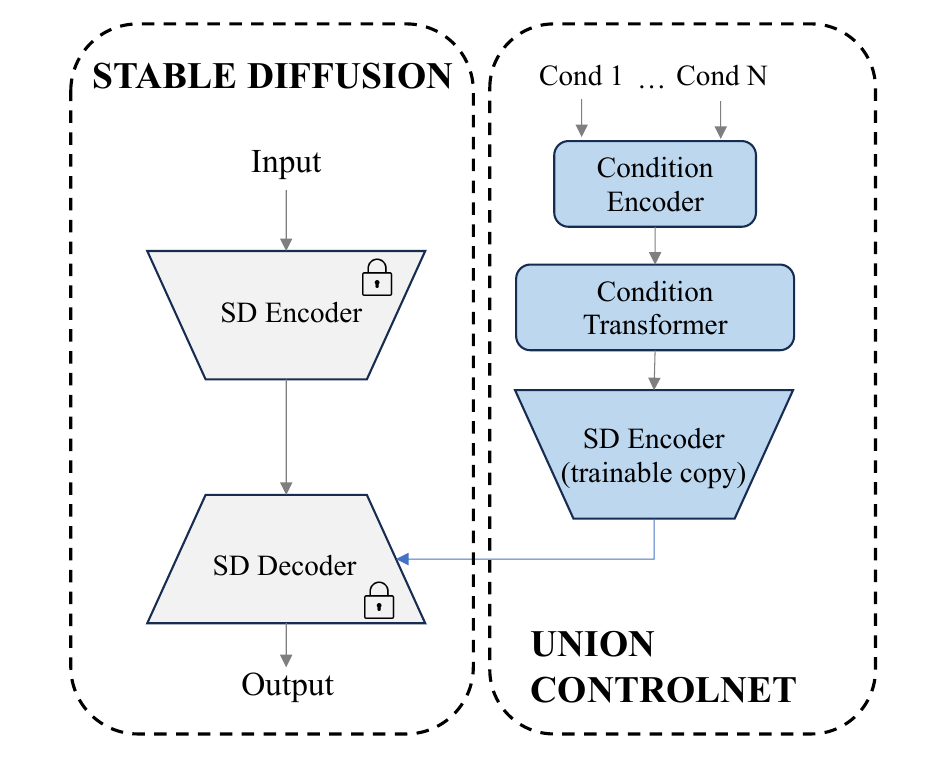} 
\caption{Architecture of union controlnet. It performs multi-source information fusion through condition encoder and transformer, enabling support for diverse types of control signals as input.} %最终文档中希望显示的图片标题
\label{fig:controlnet} %用于文内引用的标签
\end{figure}

To achieve precise alignment between the generated textures and the underlying 3D geometry, we introduce a union ControlNet\cite{controlnetplus2024} architecture that incorporates multiple geometric control signals into the generation process. This design maintains extensibility for additional control modalities, while effectively integrating diverse geometric priors to enhance structural alignment and cross-view consistency in the generated textures. The union contronet architecture is illustrated in the Fig.\ref{fig:controlnet}. Unlike the standard ControlNet\cite{zhang2023adding}, it employs a dedicated condition encoder to extract features from multiple modalities of control images. These features are then fused using a transformer module to perform cross-modal interaction. The fused representation is subsequently used as the input to the ControlNet backbone.

MVPainter incorporates two complementary types of geometric priors as control inputs: normal map and depth map. The normal map provides fine-grained local geometric details, enabling precise control over the texture generation at the micro-structural level. In contrast, the depth map offers more global spatial scale and contour information, characterized by its inherent smoothness and redundancy. This helps suppress the model’s over-reliance on noisy or misleading details in the normal map, thereby improving both stability and alignment during generation. While our current implementation leverages only normal and depth maps as control signals, the Union ControlNet architecture is inherently extensible. It supports arbitrary types and quantities of geometric inputs, such as position maps or view direction encodings, enabling flexible conditioning on multi-modal geometric information.

\noindent \textbf{Training Strategy.} Existing diffusion models are typically pretrained on large-scale image datasets, where the viewpoint distribution is heavily biased toward frontal views of objects. This bias often leads to degraded performance when generating images from non-frontal viewpoints, such as the back, sides. To address this issue, we propose a progressive three-stage training strategy that guides the model from learning overall multi-view distributions, to incorporating geometric control, and finally to fine-grained detail enhancement. This strategy progressively improves the model’s capability to synthesize high-quality results from non-frontal viewpoints.

\textbf{Stage 1: UNet Pretraining.} At this stage, we train only the UNet of the diffusion model without incorporating any geometric control signals. The objective is to enable the model to initially learn the image distribution across six target viewpoints, thereby establishing a basic multi-view generation capability.

\textbf{Stage 2: ControlNet Training.} After training the UNet, we introduce the ControlNet module and freeze the UNet weights. We then train only the ControlNet to learn the model's responsiveness to geometric control signals such as normal and depth maps. The goal of this stage is to equip the model with geometric control capabilities, enabling it to generate geometrically consistent images.

\textbf{Stage 3: Fine-tuning with High-Quality Data.} Building upon the previous two stages, we unfreeze both the UNet and ControlNet and perform joint fine-tuning using a curated set of high-quality training data as mentioned in Sec.\ref{sec:data_process}. These data samples exhibit higher texture complexity, more accurate geometric alignment, and richer detail representation, enabling the model to generate high-fidelity and detail-preserving images across all viewpoints. This stage further enhances the model’s robustness and generation quality under challenging conditions such as complex lighting and rare viewpoints.

\subsection{PBR Attributes Extractor}
In many applications that demand high levels of photorealism, generating only baked RGB texture images is insufficient. Modern graphics rendering systems typically adopt Physically-Based Rendering (PBR) to simulate the interaction between light and object surfaces. This approach requires a complete set of physical attribute maps, including basecolor (intrinsic color), metallic (degree of metalness), and roughness (surface micro-roughness). These maps respectively represent the object’s reflective color, its metallic nature, and the fine-scale texture of its surface, and are essential components for achieving realistic rendering.

To meet the above requirements, we introduce a dedicated PBR attribute extractor built upon the six-view RGB images generated in the second stage. This module takes the multi-view images $\{I_i\}_{i=1}^6$  as input and outputs basecolor, metallic, and roughness maps:
\begin{equation}
\mathcal{F}_{\text{PBR}}: \{I_i\}_{i=1}^6 \longrightarrow \{T_{\text{base}}, T_{\text{metal}}, T_{\text{rough}}\}
\end{equation}
where $\mathcal{F}_{\text{PBR}}$ denotes the PBR attribute extractor, and the outputs are basecolor map, metallic map, and roughness map, respectively.

We design our PBR attributes extractor based on the existing PBR prediction model IDArb\cite{li2024idarb} with two key improvements. First, in terms of the attention mechanism, we replace the original sequential structure of view attention, component attention, and image attention with a parallel architecture. This allows the model to perceive information from different domains at the same hierarchical level, thereby improving the efficiency of feature integration. Second, we increase the training resolution from 256×256 to 512×512. These tricks significantly enhance the PBR model’s ability to capture fine-grained details, particularly in the basecolor map.

\subsection{Implementation Details}
\noindent \textbf{Datasets.} The dataset used to train the MVPainter multi-view generation model primarily comes from \textit{Objaverse}\cite{deitke2023objaverse}. Additionally, we collected publicly available 3D models from the internet following the \textit{Objaverse} approach. Ultimately, our dataset contains approximately 1.2 million 3D models. We used the data processing pipeline outlined in Sec.\ref{sec:data_process} for rendering and scoring. For each object, we first render six images from fixed viewpoints, with azimuth and zenith angles of (0, 90, 180, 270, 0, 0) and (0, 0, 0, 0, -90, 90) degrees, respectively, along with 15 reference images conditioned on random lighting and rotations. Furthermore, we constructed a high-quality texture dataset containing 100k 3D models based on the texture scores of all objects. For training the PBR attributes extractor, we used the \textit{ARB-OBJAVERSE} dataset\cite{li2024idarb}, which includes approximately 5.7 million PBR data pairs.

\noindent \textbf{Training of Multi-view Diffusion Model.} In the first stage of training the multi-view diffusion UNet, we use the Adam optimizer with an initial learning rate of 3e-5, and apply \textit{CosineAnnealingWarmRestarts} as the learning rate scheduler. The UNet is trained on the entire 1.2 million dataset for approximately 3-5 epochs.

In the second stage, during the training of ControlNet, we fix the parameters of the UNet and employ the same learning rate, optimizer, and scheduler as in the first stage. This stage is also involved trained on the full 1.2 million dataset for about 2-3 epochs.

In the third stage, we fine-tune both ControlNet and UNet using our high-quality dataset. The initial learning rate is set to 1.5e-5, with the other parameters remaining the same as in the previous stages. This stage use 100k high-quality texture data, and the training last for 8-10 epochs.

\noindent \textbf{Training of PBR Attributes Extractor.} We train our PBR attributes extractor based on the IDArb\cite{li2024idarb}, utilizing a parallel attention mechanism. The network parameters are still initialized from the pre-trained IDArb model. Training is conducted using the \textit{ARB-OBJAVERSE} dataset, with a learning rate of 5e-6, and the\textit{ constant\_with\_warmup} learn rate scheduler. The model is trained for approximately 2-3 epochs.

\section{Experiments}
\label{sec:experiment}
In this section, we evaluate the performance of our method through a series of experiments. First, we describe the human-aligned evaluation system in Sec.\ref{sec:exp_evaluation}. Next, we compare MVPainter against several state-of-the-art baselines to assess its effectiveness in Sec.\ref{sec:exp_baseline}. Finally, we conduct an ablation study to understand the contribution of individual components in our method in Sec.\ref{sec:ablation}.

To evaluate the generalization capability of MVPainter across different objects, we employ GPT-4o\cite{hurst2024gpt} and SDXL\cite{podell2023sdxl} to generate images of common object types. First, we used GPT to generate descriptive prompts for 1,000 objects. Then, we used SDXL to generate images corresponding to these prompts, filtering out low-quality and duplicate images, resulting in a final set of 210 images. To demonstrate that MVPainter can be applied to geometries generated by various methods, we use TripoSG\cite{li2025triposg}, Hunyuan3D-2.0\cite{zhao2025hunyuan3d}, TRELLIS\cite{xiang2024structured}, and Hi3DGen\cite{ye2025hi3dgen} to generate the geometric models corresponding to these 210 images, which are then used for subsequent evaluation.

\subsection{Human-aligned Evaluation}
\label{sec:exp_evaluation}
\begin{figure}[tbp]
\centering 
\includegraphics[width=\linewidth]{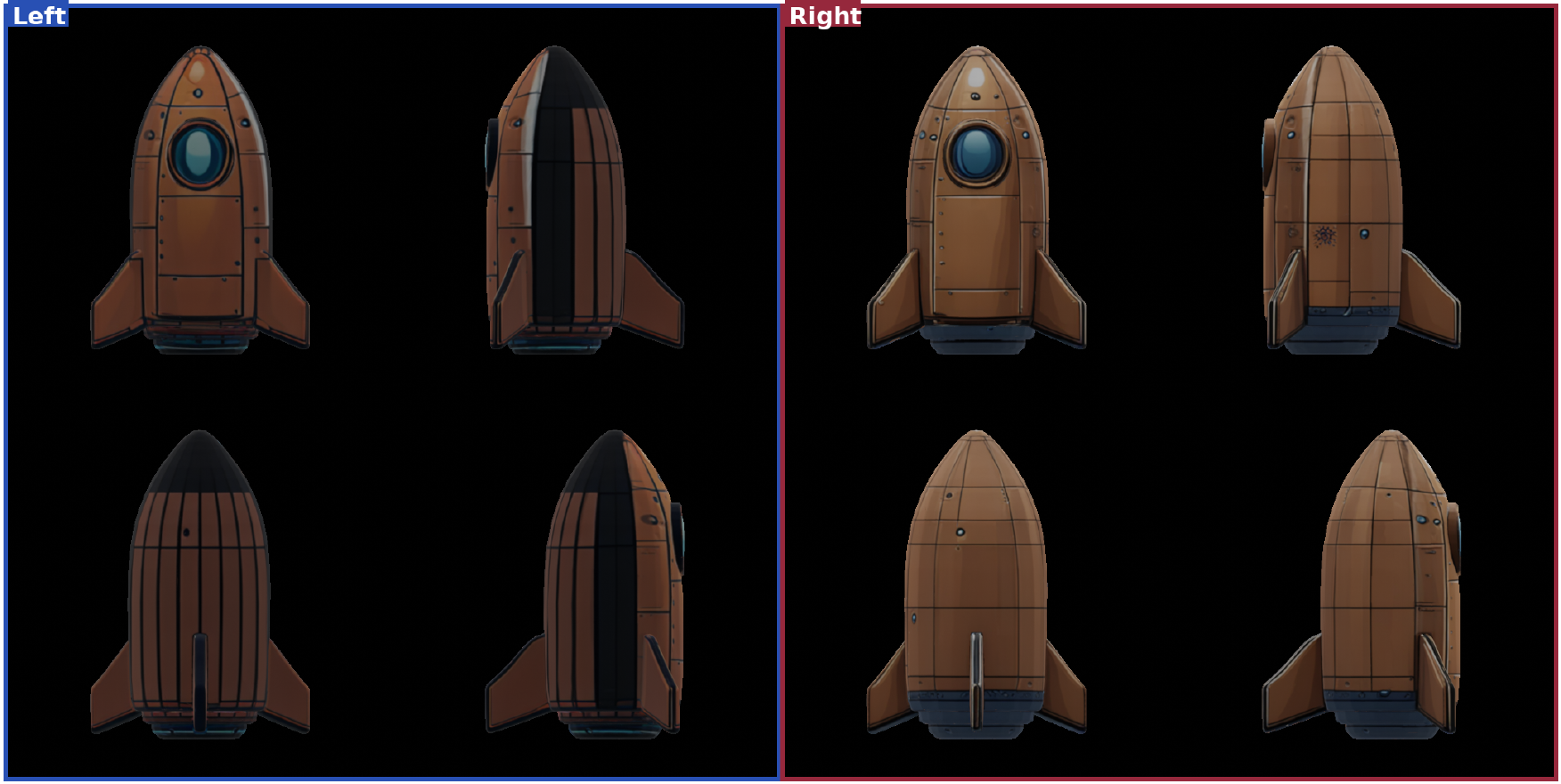} 
\caption{Example of the concatenated image which then is evaluated by VLM.} %最终文档中希望显示的图片标题
\label{fig:vlm} %用于文内引用的标签
\end{figure}

\begin{figure*}[tbp]
\centering 
\includegraphics[width=1\linewidth]{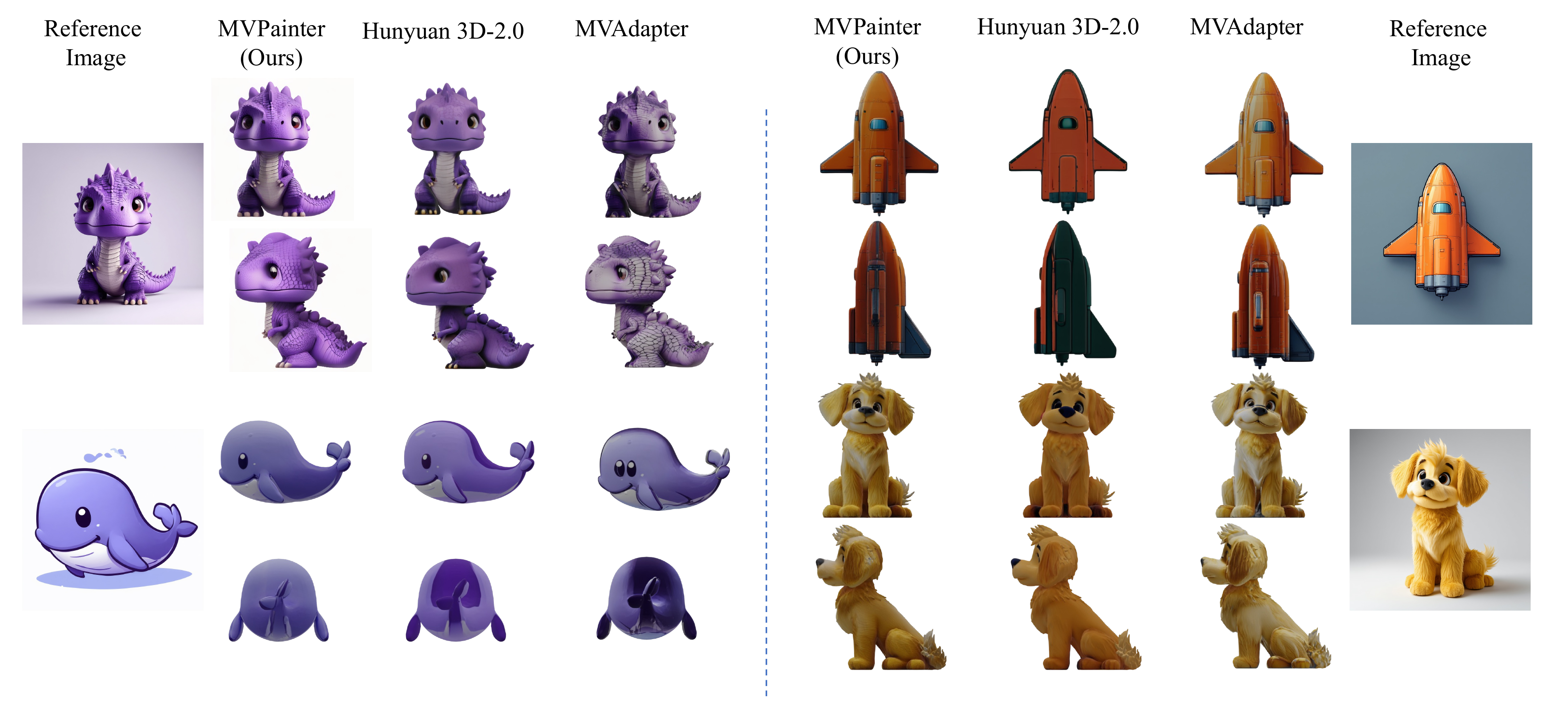} 
\caption{Qualitative comparison of baked RGB texture generation methods. Here, we only select the front and side views for illustration.} %最终文档中希望显示的图片标题
\label{fig:exp_baseline} %用于文内引用的标签
\end{figure*}

Existing 3D texture quality assessment typically employs metrics such as FID, PSNR, LPIPS, and SSIM\cite{huang2024mv,zhao2025hunyuan3d} , which rely on comparisons with ground truth. However, due to differences in image factors such as brightness and saturation, these metrics are prone to distortion and fail to accurately reflect the quality of generated textures. To address this issue, we propose a Human-Aligned evaluation system based on VLM (Visual Language Model) inspired by GPTEval3D\cite{wu2024gpt}. We design the VLM prompt for the three core dimensions of reference-texture alignment, geometry-texture consistency, and local texture quality, allowing the VLM to assess the texture quality for each respective dimension.

Assume there are \( n \) methods, denoted as \( A_1, A_2, \dots, A_n \). For each method \( i \), we render four images from their corresponding texture mesh. For any two methods \( A_i \) and \( A_j \), we concatenate their rendered images into one image \( I_{ij} \) as shown in Fig.\ref{fig:vlm}, which is then evaluated by the VLM model:

\begin{equation}
C_{ij} = \text{VLM}(I_{ij})
\end{equation}
where $C_{ij}$ is their comparison result: 
\begin{equation}
C_{ij} = \begin{cases} 
1, & \text{if } A_i \text{ is better than } A_j \\
0, & \text{if } A_j \text{ is better than } A_i \\
0.5, & \text{if } A_i \text{ and } A_j \text{ are equally ranked}
\end{cases}
\end{equation}

Then, based on these comparison results, we compute the Elo score\cite{berg2020statistical} \( R_{i} \) for each method \( A_i \) in each evaluation dimension.

Assume that our evaluation set contains \( k \) samples, then there are a total of \( \frac{n!}{2(n-2)} \times k \) pairs of comparison results. We randomly shuffle these comparison results and update the Elo score for each method using the following formula:

\begin{equation}
R_{i} = R_{i} + K \cdot \left( C_{ij} - E_{ij} \right)
\end{equation}

where $K$ is a constant factor for Elo adjustment. $E_{ij}$ is the expected score based on the current Elo ratings of $A_i$ and $A_j$:
\begin{equation}
E_{ij} = \frac{1}{1 + 10^{(R_{j} - R_{i})/{400}}}
\end{equation}

To avoid the errors caused by order dependency in the Elo scores, we randomly shuffle the comparison results and repeat the Elo score calculation 100 times. Finally, we take the average of the 100 computed results as the final Elo score for each method.

It is noteworthy that we do not concatenate the results of all methods for direct comparison using VLM. The reason is we observed that when excessively large images are concatenated, the VLM( we use \textit{QWen2.5-VL-32B}\cite{yang2024qwen2} as our evaluation VLM model) struggles to capture local details, which increases the likelihood of inaccurate evaluation results. To mitigate this issue, we employ a pairwise comparison approach for all methods, thereby ensuring the accuracy of the VLM evaluation. Ultimately, the pairwise comparison results are quantified using the Elo rating.

\subsection{Comparison with Baselines}
\label{sec:exp_baseline}
MVPainter is capable of generating RGB textures that are consistent with the reference image, as well as further extracting PBR textures from these generated RGB textures. For RGB texture generation, we compared MVPainter with existing state-of-the-art open-source methods Hunyuan3D-2.0\cite{zhao2025hunyuan3d} and MVAdapter\cite{huang2024mv}. For PBR texture generation, we compared MVPainter with the baseline method, IDArb\cite{li2024idarb}.

\begin{figure*}[htbp]
\centering 
\includegraphics[width=\linewidth]{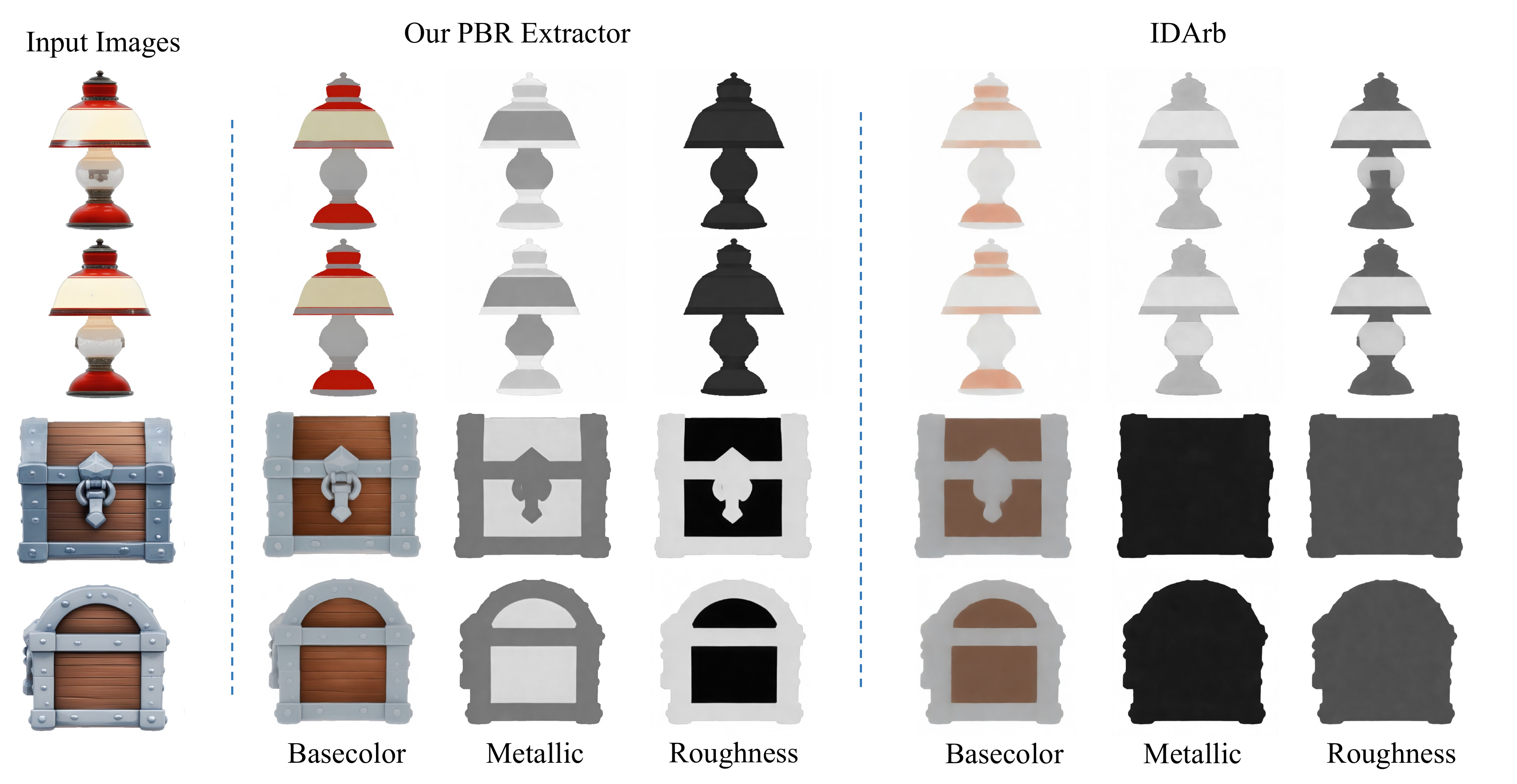} 
\caption{Qualitative comparison of our PBR extractor and its baseline IDArb. Our model can capture more accurate details than IDArb.} %最终文档中希望显示的图片标题
\label{fig:exp_pbr} %用于文内引用的标签
\end{figure*}

\begin{figure*}[htbp]
\centering 
\includegraphics[width=\linewidth]{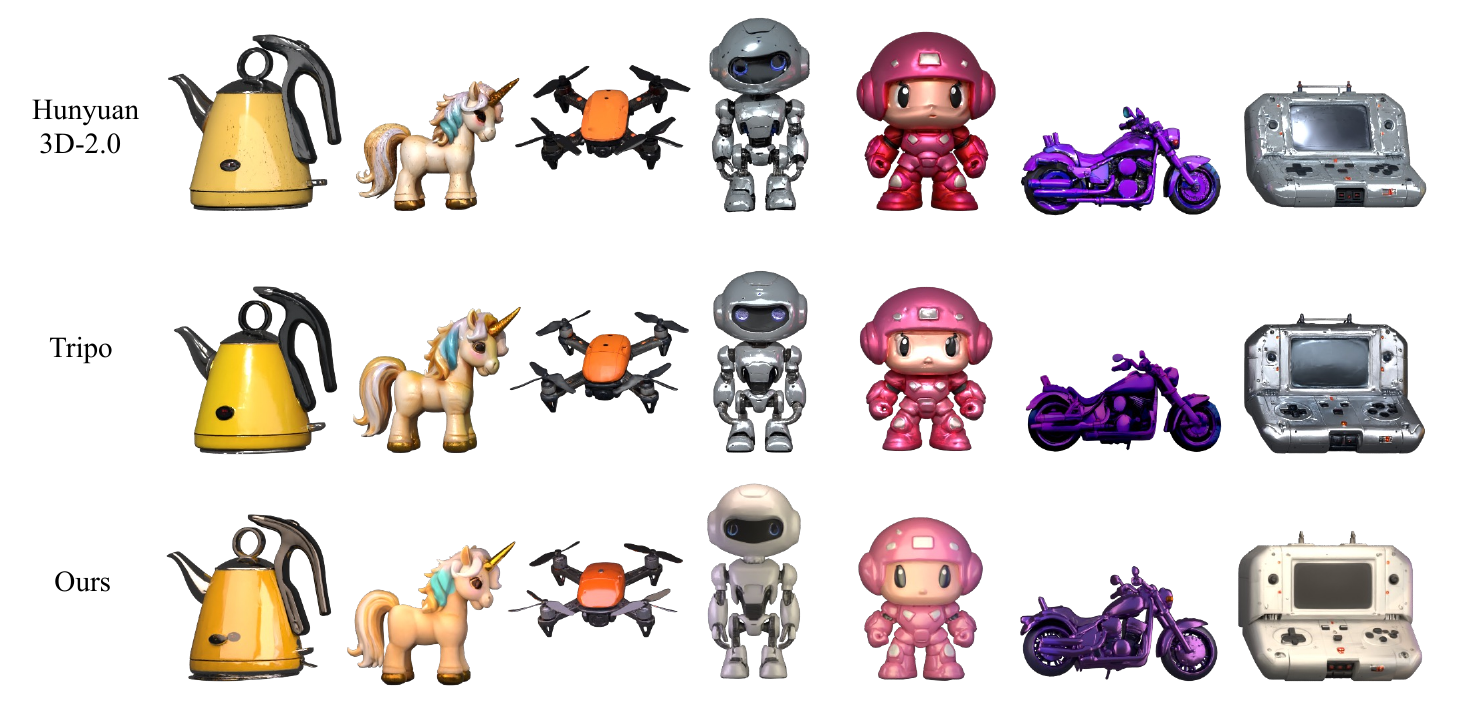} 
\caption{Our PBR extractor can achieve comparable performance to commercial applications.} %最终文档中希望显示的图片标题
\label{fig:exp_more_pbr} %用于文内引用的标签
\captionsetup{skip=0pt}

\end{figure*}

\begin{figure*}[!p]
\centering 
\includegraphics[width=0.85\linewidth]{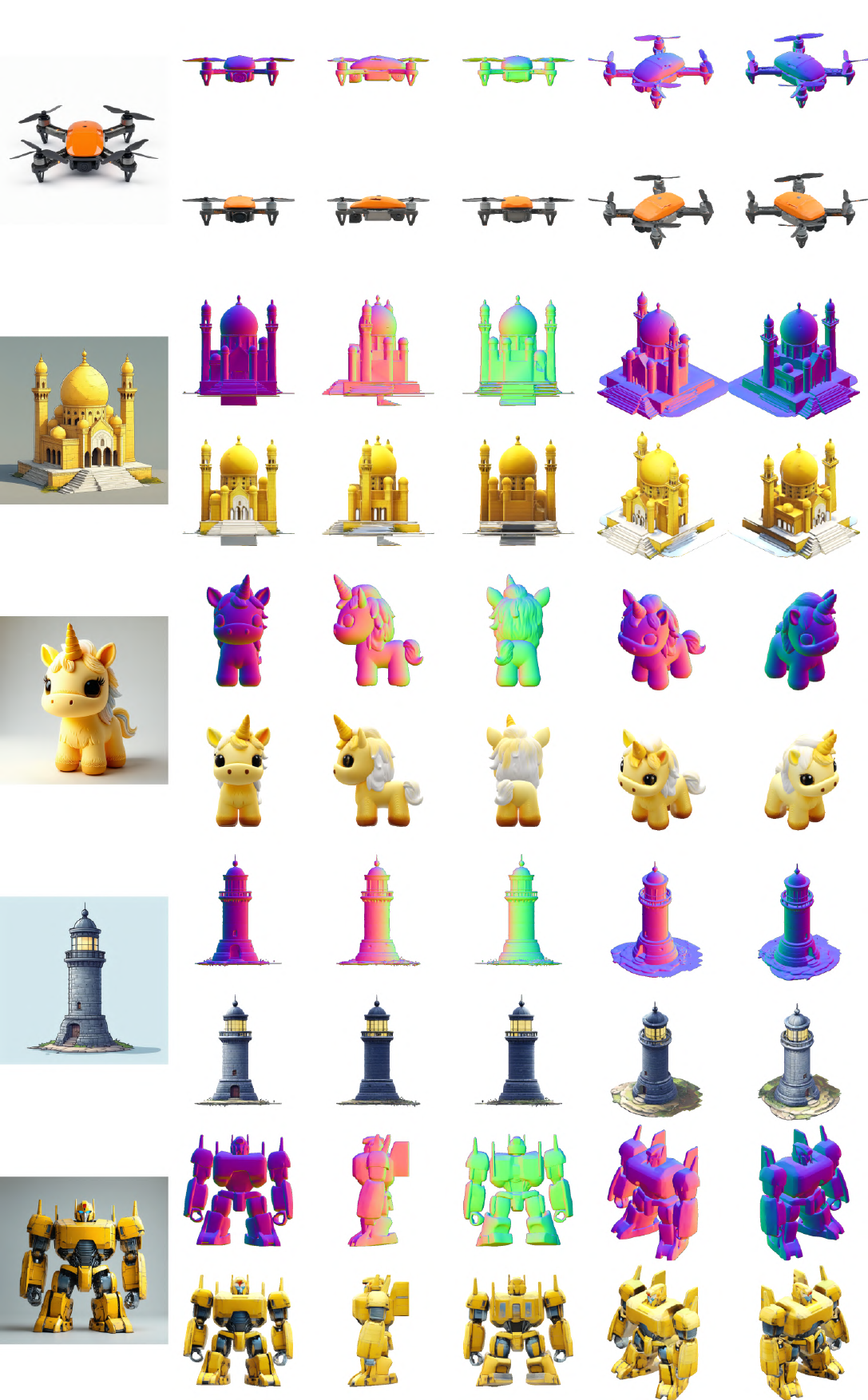} 
\caption{MVPainter's performance on the geometries generated by TripoSG. The first column is the reference image, and the first row of each object is the corresponding normal map.} %最终文档中希望显示的图片标题
\label{fig:result_tripo} %用于文内引用的标签
\end{figure*}

\begin{figure*}[!p]
\centering 
\includegraphics[width=0.85\linewidth]{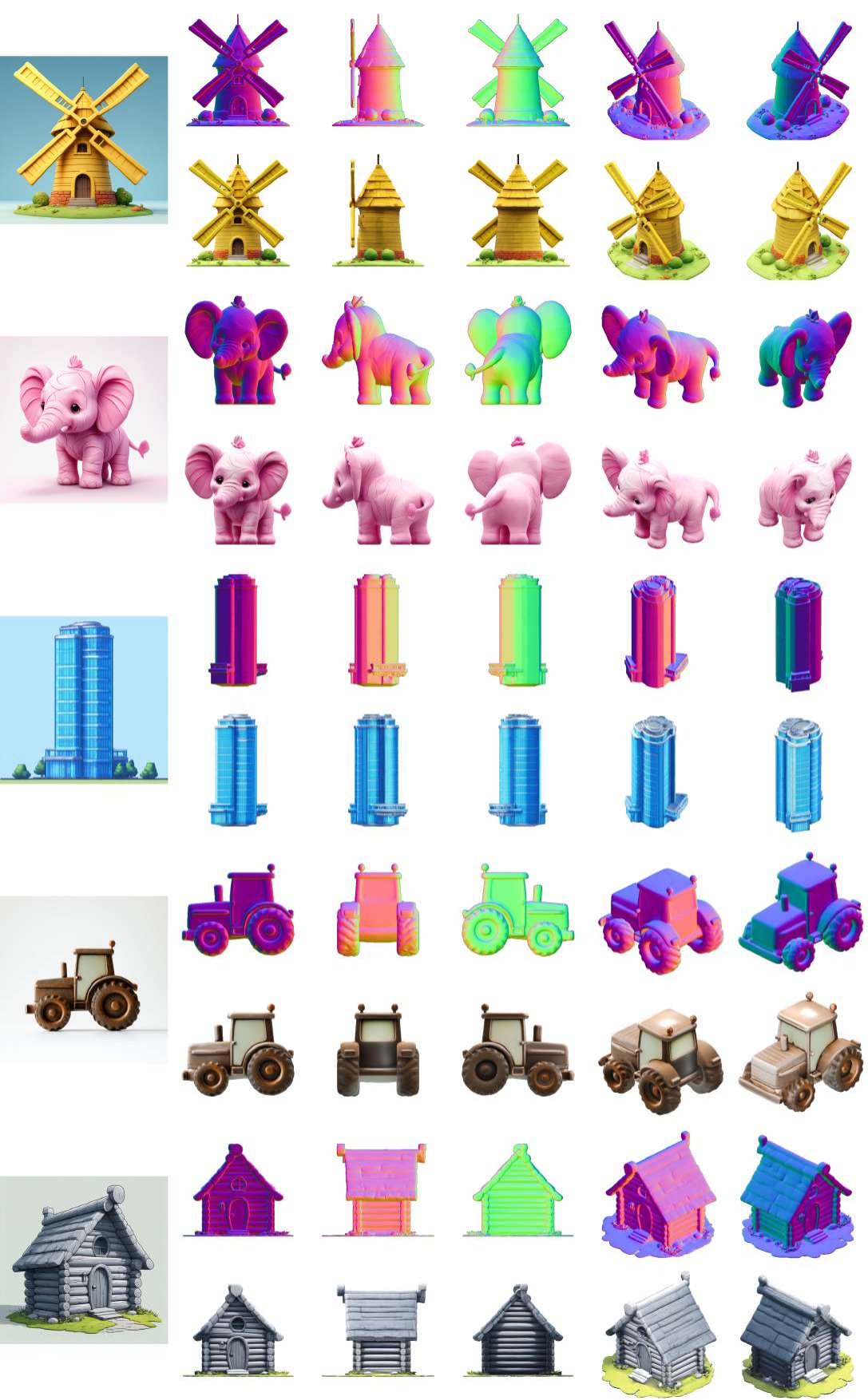} 
\caption{MVPainter's performance on the geometries generated by Hunyuan3D-2.0.} %最终文档中希望显示的图片标题
\label{fig:result_hunyuan} %用于文内引用的标签
\end{figure*}

\begin{figure*}[!p]
\centering 
\includegraphics[width=0.85\linewidth]{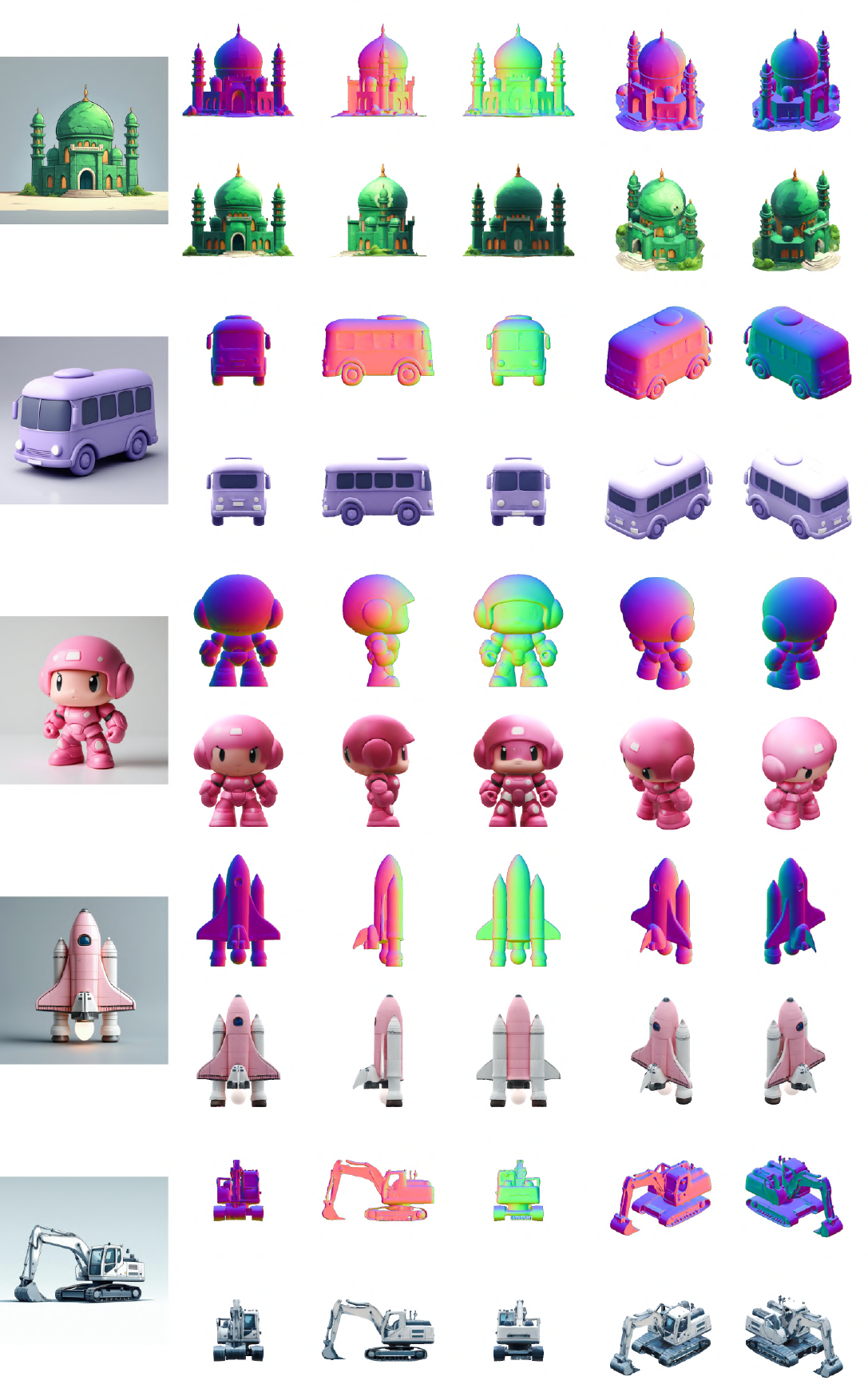} 
\caption{MVPainter's performance on the geometries generated by TRELLIS.} %最终文档中希望显示的图片标题
\label{fig:result_trellis} %用于文内引用的标签
\end{figure*}

\begin{figure*}[!p]
\centering 
\includegraphics[width=0.85\linewidth]{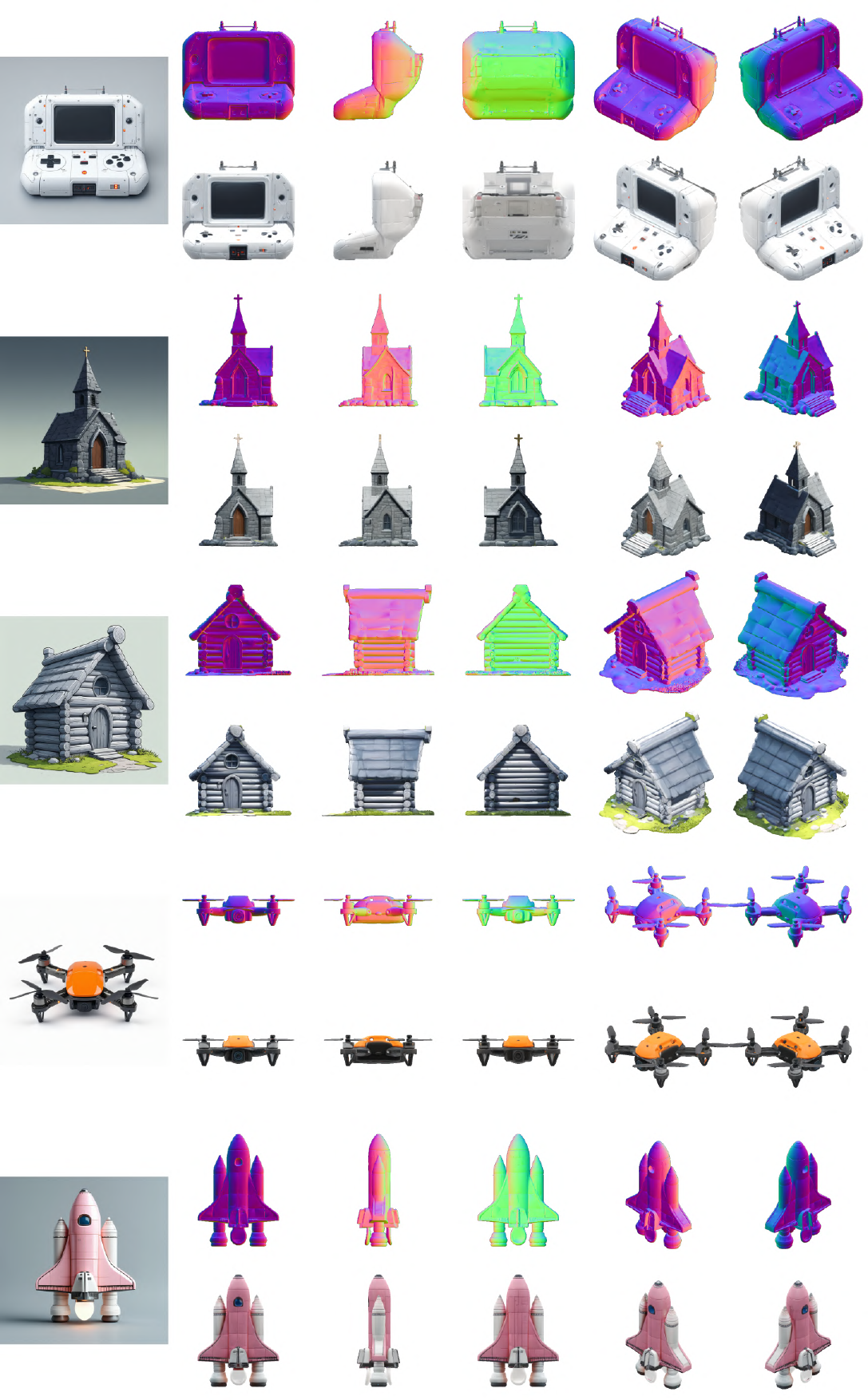} 
\caption{MVPainter's performance on the geometries generated by Hi3DGen.} %最终文档中希望显示的图片标题
\label{fig:result_hi3d} %用于文内引用的标签
\end{figure*}

We evaluate MVPainter with two 3D texture generation methods: Hunyuan3D-2.0, and MVAdapter. Fig.\ref{fig:exp_baseline} presents the qualitative results of these methods across four distinct cases. Hunyuan3D-2.0 benefits from its Delighting module, effectively removing lighting from the reference images, resulting in clean textures with consistent color styles. However, this method has the drawback of potentially altering the object's intrinsic color, as seen in the fourth dog case, and it generally lacks fine details, as many of the details in the reference image are enhanced through highlights. When these highlights are removed, it is difficult for the diffusion model to capture these details. MVAdapter generates textures with rich details but suffers from poor alignment with geometry. For example, in the whale case, the eyes are incorrectly positioned, and in the dog case, the nose and eyes show slight misalignment. Additionally, MVAdapter's handling of lighting is suboptimal; in the first case, it mistakenly interprets highlights as texture, leading to odd texture results. In contrast, our method MVPainter, with its carefully designed data processing, augmentation techniques, and model architecture, performs better in handling lighting, geometric consistency, and local detail preservation. This conclusion is also supported by the quantitative experiments in Table.\ref{tab:elo}, which are the Elo scores calculated by different texture generation methods using the VLM-based approach mentioned in Sec.\ref{sec:exp_evaluation}. The quantitative results show that MVPainter achieves the best performance across geometries generated by different methods, highlighting its robustness to various geometric inputs. More visualization results of MVPainter can be found in Fig.\ref{fig:result_tripo}, \ref{fig:result_hunyuan}, \ref{fig:result_trellis}, \ref{fig:result_hi3d} at the end of the report.

\begin{table*}[htbp]
  \centering
  \caption{Elo scores of different 3D texture generation methods under various geometry methods.}
    \resizebox{0.8\linewidth}{!}{

    \begin{tabular}{c|cccc}
    \toprule
    \multicolumn{1}{l|}{Geometry} & Method & \multicolumn{1}{l}{\makecell{Reference-texture \\Alignment}} & \multicolumn{1}{l}{\makecell{Geometry-texture\\ Consistency}} & \multicolumn{1}{l}{\makecell{Local Texture \\Quality }} \\
    \midrule
    \multirow{3}[2]{*}{TripoSG} & Hunyuan2 & 816   & 838   & 816   \\
          & MVAdapter & 1030  & 1026  & 1050   \\
          & \textbf{MVPainter(Ours)} & 1109  & 1136  & 1134  \\
    \midrule
    \multirow{3}[2]{*}{Hunyuan2} & Hunyuan2 & 846   & 835   & 824   \\
          & MVAdapter & 1034  & 1032  & 1031   \\
          & \textbf{MVPainter(Ours)} & 1120  & 1132  & 1146   \\
    \midrule
    \multirow{3}[2]{*}{TRELLIS} & Hunyuan2 & 861   & 846   & 828    \\
          & MVAdapter & 1033  & 1040  & 1056  \\
          & \textbf{MVPainter(Ours)} & 1105  & 1114  & 1117  \\
    \midrule
    \multirow{3}[2]{*}{Hi3dGen} & Hunyuan2 & 865   & 859   & 837    \\
          & MVAdapter & 1010  & 993   & 1008 \\
          & \textbf{MVPainter(Ours)} & 1125  & 1148  & 1156 \\
    \bottomrule
    \end{tabular}}
  \label{tab:elo}%
\end{table*}%

\begin{table*}[htbp]
  \centering
  \caption{Average Elo scores calculated based on five human evaluations.}
  \resizebox{0.7\linewidth}{!}{
    \begin{tabular}{cccc}
    \toprule
    Method & \multicolumn{1}{l}{\makecell{Reference-texture \\Alignment}} & \multicolumn{1}{l}{\makecell{Geometry-texture\\ Consistency}} & \multicolumn{1}{l}{\makecell{Local Texture \\Quality }}  \\
    \midrule
    Hunyuan2 & 918   & 628   & 845 \\
    MVAdapter & 938   & 1180  & 1041 \\
    \textbf{MVPainter(Ours)} & 1144  & 1192  & 1114 \\
    \bottomrule
    \end{tabular}%
    }
  \label{tab:elo_human}%
\end{table*}%

To demonstrate that our evaluation method in Sec.\ref{sec:exp_evaluation} is aligned with human judgment, we conduct a user study on the textured TripoSG meshes. We recruit five human participants and ask them to perform pairwise comparisons. We then compute Elo scores using the same procedure described in Sec. \ref{sec:exp_evaluation}. The results are summarized in Table.\ref{tab:elo_human}. As shown, the trends in human evaluation closely match those obtained by the VLM-based evaluation, though there are some differences in the exact Elo score margins. Notably, discrepancies are more pronounced in dimensions that require detailed visual inspection, such as \textit{Geometry-Texture Consistency} and \textit{Local Texture Quality}, where current VLMs still fall short of capturing fine-grained details as reliably as humans. Nonetheless, VLMs have proven to be sufficiently reliable for assessing texture quality at a global level. We believe that with ongoing advancements in VLM technology, their ability to capture fine details will increasingly approach human-level accuracy.

Fig.\ref{fig:exp_pbr} presents a comparison between our PBR attributes extractor and its baseline model, IDArb. Compared to IDArb, we have adopted a parallelized attention mechanism in our model, which includes component attention, view attention, and image cross-attention. This improvement helps better preserve the domain information of each attention operation and enhances the efficiency of information exchange between modules. Furthermore, we have increased the training resolution from 256 to 512 to improve the model's ability to capture and generate fine details. Qualitative evaluation results indicate that our improvements significantly enhance IDArb’s performance in extracting PBR textures from multi-view images. Specifically, the modified model effectively removes lighting effects while preserving more surface details of the object in basecolor extraction, benefiting the PBR textures with superior performance in both visual fidelity and geometric consistency. We futher compare out PBR extractor with Tripo and Hunyuan 3D-2.0, the PBR models of Tripo and Hunyuan3D-2.0 are generated from their official website applications, their pbr generation models are currently closed source. Fig.\ref{fig:exp_pbr} shows that our PBR model can achieve comparable performance to commercial applications.

\subsection{Ablation Study}
\label{sec:ablation}
\begin{figure}[htbp]
\centering 
\includegraphics[width=0.9\linewidth]{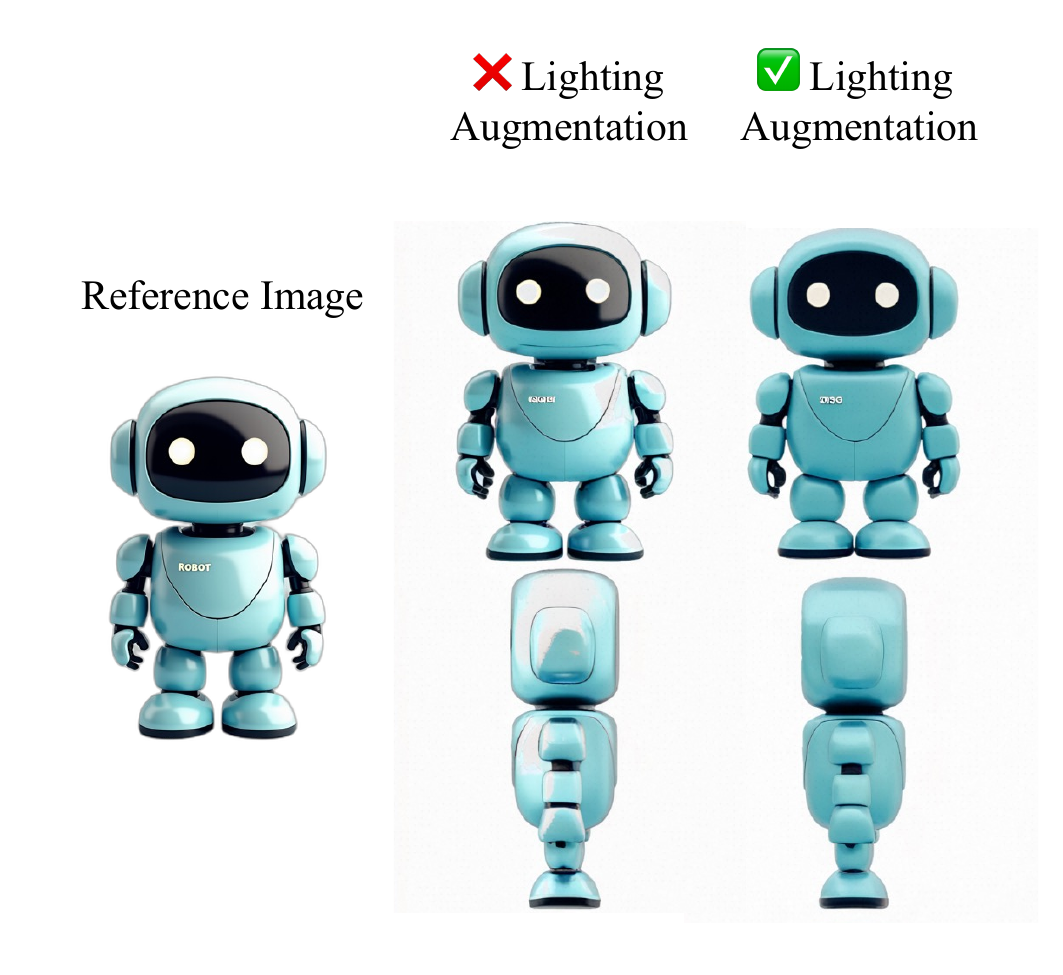} 
\caption{Importance of the lighting augmentation in our data process pipeline. The images generated after adding lighting augmentation can avoid the highlight effect in the reference image.} %最终文档中希望显示的图片标题
\label{fig:ablation_light} %用于文内引用的标签
\end{figure}

\begin{figure}[htbp]
\centering 
\includegraphics[width=\linewidth]{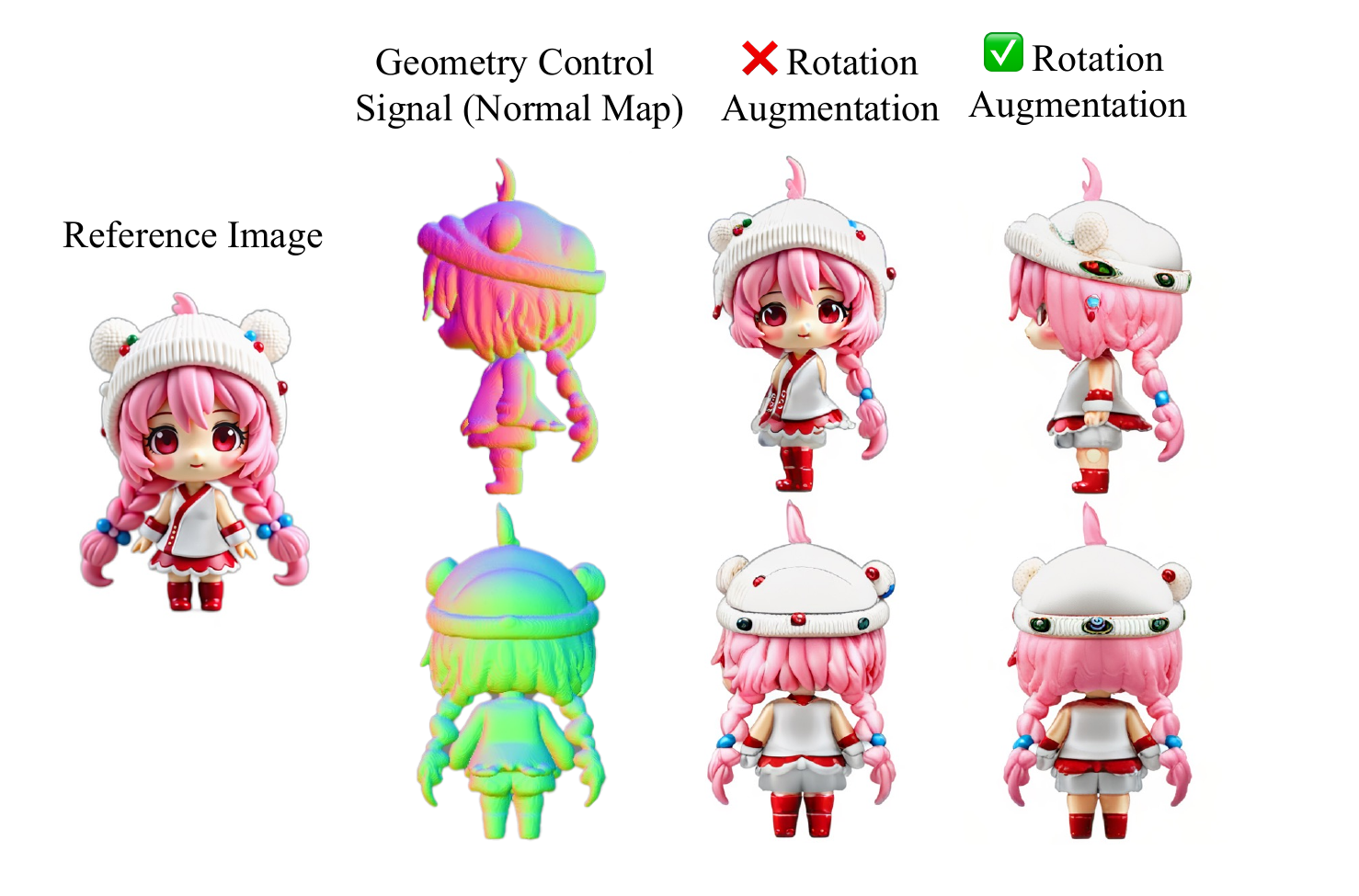} 
\caption{Importance of the rotation augmentation. The model after adding the rotation enhancement has a strong generalization ability for non-frontal perspectives} %最终文档中希望显示的图片标题
\label{fig:ablation_rotation} %用于文内引用的标签
\end{figure}

\noindent\textbf{Importance of Lighting/Rotation Augmentation}. As described in Sec.\ref{sec:data_process}, we employ lighting and rotation data augmentation to simulate the significant variations in lighting and rotation found in the in-the-wild inputs. We find that both augmentation techniques are essential and provide a notable improvement in the model's generalization capability. Fig.\ref{fig:ablation_light} and Fig.\ref{fig:ablation_rotation}present the ablation study results for these data augmentations. Without lighting augmentation, the diffusion model tends to bake all lighting effects into the texture maps. Although the generated textures appear more "realistic," when the object is placed in an interactive environment, this baked lighting effect reduces the model's realism. For rotation augmentation, we test by rotating the geometry aligned with the reference image by 90 degrees as control signals. It can be seen that, without rotation augmentation, the first texture generated by the diffusion model is severely misaligned with the geometry. The reason for this phenomenon is that the target image from the first perspective in the training set is highly similar to the reference image, prompting the diffusion model to rely directly on the reference image rather than inferring from the geometric control information. Rotation augmentation prevents this "shortcut" behavior in the diffusion model training. As proven by the ablation study above, both lighting and rotation augmentations help mitigate the dependency on reference images, enhancing MVPainter's generalization ability when faced with reference images outside the training set distribution.


\begin{figure}[htbp]
\centering 
\includegraphics[width=\linewidth]{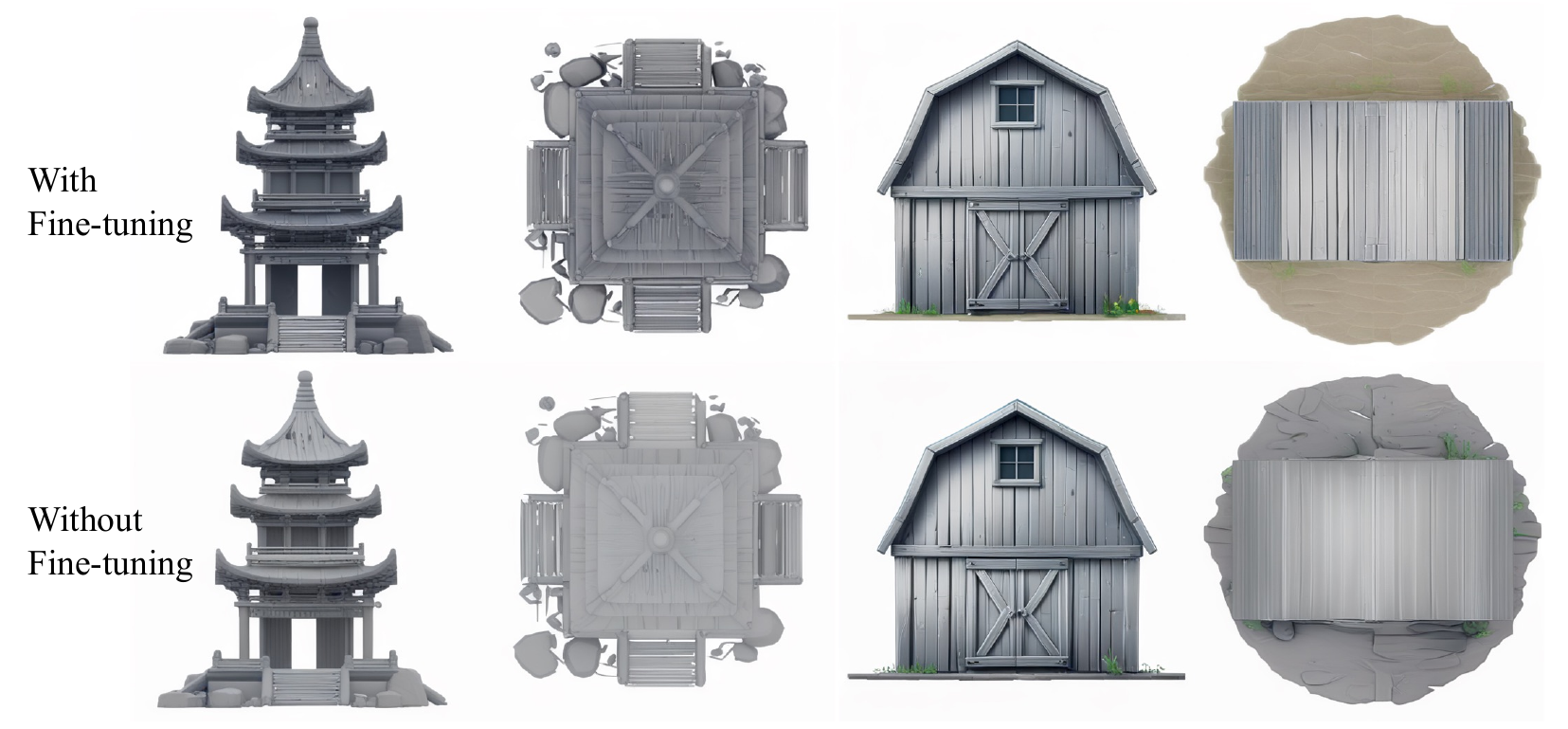} 
\caption{Importance of fine-tuning with high-quality dataset. After fine-tuning with high-quality dataset, MVPainter is able to produce richer and better local details.} %最终文档中希望显示的图片标题
\label{fig:ablation_finetune} %用于文内引用的标签
\end{figure}

\noindent\textbf{Importance of High-quality Fine-tuning.}
As mentioned in the Sec.\ref{sec:multi_view_generation}, after training ControlNet, we fine-tuned it using a high-quality dataset. The purpose of this fine-tuning is to enable the diffusion model to learn the rich texture details present in the high-quality dataset, thereby improving the generated local texture quality. We visualize the effect of this high-quality fine-tuning phase, as shown in Fig.\ref{fig:ablation_finetune}. After fine-tuning with the high-quality dataset, the generated textures exhibit significantly richer local details, particularly from the top perspective. This is because the top view, compared to directly extracting information from the reference image, benefits more directly from the rich texture distribution in the high-quality dataset.

\section{Conclusion}
In this report, we thoroughly explore the three core challenges in 3D texture generation and propose a systematic solution. By constructing a high-quality data process pipeline, introducing a ControlNet-based geometric control structure, and designing a VLM-based human perception-aligned evaluation strategy, we successfully enhanced and evaluated MVPainter's performance in three key areas: reference-texture alignment, geometry-texture consistency, and local texture quality. Furthermore, we have open-sourced MVPainter, a complete PBR texture generation system, including data construction tools, training frameworks, and evaluation scripts. We hope MVPainter will serve as a valuable tool for the research community, advancing the standardization, reproducibility, and scalability of 3D texture generation tasks and contributing to the progress of 3D generation research.

Although MVPainter has made progress in 3D texture generation, there are still many issues that need further exploration. The current MVPainter system only supports the generation of a fixed set of six viewpoints, which is sufficient for most objects. However, for objects with self-occlusion, complete coverage remains challenging. Therefore, generating textures from variable and extreme viewpoints is an issue that needs to be addressed. Moreover, the MVPainter system has not yet been fully optimized for speed. The computational efficiency still has room for improvement, especially when dealing with large-scale data or real-time generation requirements. Therefore, improving generation speed and reducing computational costs will be a key focus for our future optimization efforts.

\newpage
{
    \small
    \bibliographystyle{ieeenat_fullname}
    \bibliography{main}

\begin{thebibliography}{28}
\providecommand{\natexlab}[1]{#1}
\providecommand{\url}[1]{\texttt{#1}}
\expandafter\ifx\csname urlstyle\endcsname\relax
  \providecommand{\doi}[1]{doi: #1}\else
  \providecommand{\doi}{doi: \begingroup \urlstyle{rm}\Url}\fi

\bibitem[Bensadoun et~al.(2024)Bensadoun, Monnier, Kleiman, Kokkinos, Siddiqui, Kariya, Harosh, Shapovalov, Graham, Garreau, et~al.]{bensadoun2024meta}
Raphael Bensadoun, Tom Monnier, Yanir Kleiman, Filippos Kokkinos, Yawar Siddiqui, Mahendra Kariya, Omri Harosh, Roman Shapovalov, Benjamin Graham, Emilien Garreau, et~al.
\newblock Meta 3d gen.
\newblock \emph{arXiv preprint arXiv:2407.02599}, 2024.

\bibitem[Berg(2020)]{berg2020statistical}
Arthur Berg.
\newblock Statistical analysis of the elo rating system in chess.
\newblock \emph{Chance}, 33\penalty0 (3):\penalty0 31--38, 2020.

\bibitem[Cao et~al.(2024)Cao, Cao, Han, Shan, and Wong]{cao2024dreamavatar}
Yukang Cao, Yan-Pei Cao, Kai Han, Ying Shan, and Kwan-Yee~K Wong.
\newblock Dreamavatar: Text-and-shape guided 3d human avatar generation via diffusion models.
\newblock In \emph{Proceedings of the IEEE/CVF Conference on Computer Vision and Pattern Recognition}, pages 958--968, 2024.

\bibitem[Deitke et~al.(2023)Deitke, Schwenk, Salvador, Weihs, Michel, VanderBilt, Schmidt, Ehsani, Kembhavi, and Farhadi]{deitke2023objaverse}
Matt Deitke, Dustin Schwenk, Jordi Salvador, Luca Weihs, Oscar Michel, Eli VanderBilt, Ludwig Schmidt, Kiana Ehsani, Aniruddha Kembhavi, and Ali Farhadi.
\newblock Objaverse: A universe of annotated 3d objects.
\newblock In \emph{Proceedings of the IEEE/CVF conference on computer vision and pattern recognition}, pages 13142--13153, 2023.

\bibitem[Guo et~al.(2020)Guo, Zuo, Wang, Zou, Sun, Deng, Gong, and Cheng]{guo2020action2motion}
Chuan Guo, Xinxin Zuo, Sen Wang, Shihao Zou, Qingyao Sun, Annan Deng, Minglun Gong, and Li Cheng.
\newblock Action2motion: Conditioned generation of 3d human motions.
\newblock In \emph{Proceedings of the 28th ACM International Conference on Multimedia}, pages 2021--2029, 2020.

\bibitem[Hu et~al.(2024)Hu, Zhao, and Liu]{hu2024game}
Chengpeng Hu, Yunlong Zhao, and Jialin Liu.
\newblock Game generation via large language models.
\newblock In \emph{2024 IEEE Conference on Games (CoG)}, pages 1--4. IEEE, 2024.

\bibitem[Huang et~al.(2024{\natexlab{a}})Huang, Shao, Zhang, Zhang, Feng, Liu, and Wang]{huang2024humannorm}
Xin Huang, Ruizhi Shao, Qi Zhang, Hongwen Zhang, Ying Feng, Yebin Liu, and Qing Wang.
\newblock Humannorm: Learning normal diffusion model for high-quality and realistic 3d human generation.
\newblock In \emph{Proceedings of the IEEE/CVF Conference on Computer Vision and Pattern Recognition}, pages 4568--4577, 2024{\natexlab{a}}.

\bibitem[Huang et~al.(2024{\natexlab{b}})Huang, Guo, Wang, Yi, Ma, Cao, and Sheng]{huang2024mv}
Zehuan Huang, Yuan-Chen Guo, Haoran Wang, Ran Yi, Lizhuang Ma, Yan-Pei Cao, and Lu Sheng.
\newblock Mv-adapter: Multi-view consistent image generation made easy.
\newblock \emph{arXiv preprint arXiv:2412.03632}, 2024{\natexlab{b}}.

\bibitem[Hurst et~al.(2024)Hurst, Lerer, Goucher, Perelman, Ramesh, Clark, Ostrow, Welihinda, Hayes, Radford, et~al.]{hurst2024gpt}
Aaron Hurst, Adam Lerer, Adam~P Goucher, Adam Perelman, Aditya Ramesh, Aidan Clark, AJ Ostrow, Akila Welihinda, Alan Hayes, Alec Radford, et~al.
\newblock Gpt-4o system card.
\newblock \emph{arXiv preprint arXiv:2410.21276}, 2024.

\bibitem[Jiang et~al.(2024)Jiang, Yu, Xie, Li, Feng, Wang, Li, Lau, Gao, Yang, et~al.]{jiang2024vr}
Ying Jiang, Chang Yu, Tianyi Xie, Xuan Li, Yutao Feng, Huamin Wang, Minchen Li, Henry Lau, Feng Gao, Yin Yang, et~al.
\newblock Vr-gs: A physical dynamics-aware interactive gaussian splatting system in virtual reality.
\newblock In \emph{ACM SIGGRAPH 2024 Conference Papers}, pages 1--1, 2024.

\bibitem[Li et~al.(2024{\natexlab{a}})Li, Liu, Chen, Liang, Chen, Tan, and Long]{li2024craftsman}
Weiyu Li, Jiarui Liu, Rui Chen, Yixun Liang, Xuelin Chen, Ping Tan, and Xiaoxiao Long.
\newblock Craftsman: High-fidelity mesh generation with 3d native generation and interactive geometry refiner.
\newblock \emph{arXiv preprint arXiv:2405.14979}, 2024{\natexlab{a}}.

\bibitem[Li et~al.(2024{\natexlab{b}})Li, Zhang, Kang, Cheng, Gao, Zhang, Liang, Liao, Cao, and Shan]{li2024advances}
Xiaoyu Li, Qi Zhang, Di Kang, Weihao Cheng, Yiming Gao, Jingbo Zhang, Zhihao Liang, Jing Liao, Yan-Pei Cao, and Ying Shan.
\newblock Advances in 3d generation: A survey.
\newblock \emph{arXiv preprint arXiv:2401.17807}, 2024{\natexlab{b}}.

\bibitem[Li et~al.(2025)Li, Zou, Liu, Wang, Liang, Yu, Liu, Guo, Liang, Ouyang, et~al.]{li2025triposg}
Yangguang Li, Zi-Xin Zou, Zexiang Liu, Dehu Wang, Yuan Liang, Zhipeng Yu, Xingchao Liu, Yuan-Chen Guo, Ding Liang, Wanli Ouyang, et~al.
\newblock Triposg: High-fidelity 3d shape synthesis using large-scale rectified flow models.
\newblock \emph{arXiv preprint arXiv:2502.06608}, 2025.

\bibitem[Li et~al.(2024{\natexlab{c}})Li, Wu, Tan, Zhang, Wang, and Lin]{li2024idarb}
Zhibing Li, Tong Wu, Jing Tan, Mengchen Zhang, Jiaqi Wang, and Dahua Lin.
\newblock Idarb: Intrinsic decomposition for arbitrary number of input views and illuminations.
\newblock \emph{arXiv preprint arXiv:2412.12083}, 2024{\natexlab{c}}.

\bibitem[Podell et~al.(2023)Podell, English, Lacey, Blattmann, Dockhorn, M{\"u}ller, Penna, and Rombach]{podell2023sdxl}
Dustin Podell, Zion English, Kyle Lacey, Andreas Blattmann, Tim Dockhorn, Jonas M{\"u}ller, Joe Penna, and Robin Rombach.
\newblock Sdxl: Improving latent diffusion models for high-resolution image synthesis.
\newblock \emph{arXiv preprint arXiv:2307.01952}, 2023.

\bibitem[Shi et~al.(2023)Shi, Chen, Zhang, Liu, Xu, Wei, Chen, Zeng, and Su]{shi2023zero123++}
Ruoxi Shi, Hansheng Chen, Zhuoyang Zhang, Minghua Liu, Chao Xu, Xinyue Wei, Linghao Chen, Chong Zeng, and Hao Su.
\newblock Zero123++: a single image to consistent multi-view diffusion base model.
\newblock \emph{arXiv preprint arXiv:2310.15110}, 2023.

\bibitem[Wei et~al.(2025)Wei, Wang, Zhou, Chen, and Wang]{wei2025octgpt}
Si-Tong Wei, Rui-Huan Wang, Chuan-Zhi Zhou, Baoquan Chen, and Peng-Shuai Wang.
\newblock Octgpt: Octree-based multiscale autoregressive models for 3d shape generation.
\newblock \emph{arXiv preprint arXiv:2504.09975}, 2025.

\bibitem[Werning(2024)]{werning2024generative}
Stefan Werning.
\newblock Generative ai and the technological imaginary of game design.
\newblock In \emph{Creative Tools and the Softwarization of Cultural Production}, pages 67--90. Springer, 2024.

\bibitem[Wu et~al.(2024)Wu, Yang, Li, Zhang, Liu, Guibas, Lin, and Wetzstein]{wu2024gpt}
Tong Wu, Guandao Yang, Zhibing Li, Kai Zhang, Ziwei Liu, Leonidas Guibas, Dahua Lin, and Gordon Wetzstein.
\newblock Gpt-4v (ision) is a human-aligned evaluator for text-to-3d generation.
\newblock In \emph{Proceedings of the IEEE/CVF conference on computer vision and pattern recognition}, pages 22227--22238, 2024.

\bibitem[Xiang et~al.(2024)Xiang, Lv, Xu, Deng, Wang, Zhang, Chen, Tong, and Yang]{xiang2024structured}
Jianfeng Xiang, Zelong Lv, Sicheng Xu, Yu Deng, Ruicheng Wang, Bowen Zhang, Dong Chen, Xin Tong, and Jiaolong Yang.
\newblock Structured 3d latents for scalable and versatile 3d generation.
\newblock \emph{arXiv preprint arXiv:2412.01506}, 2024.

\bibitem[xinsir6(2024)]{controlnetplus2024}
xinsir6.
\newblock Controlnet++: All-in-one controlnet for image generations and editing!
\newblock \url{https://github.com/xinsir6/ControlNetPlus}, 2024.

\bibitem[Yang et~al.(2024)Yang, Yang, Zhang, Hui, Zheng, Yu, Li, Liu, Huang, Wei, et~al.]{yang2024qwen2}
An Yang, Baosong Yang, Beichen Zhang, Binyuan Hui, Bo Zheng, Bowen Yu, Chengyuan Li, Dayiheng Liu, Fei Huang, Haoran Wei, et~al.
\newblock Qwen2. 5 technical report.
\newblock \emph{arXiv preprint arXiv:2412.15115}, 2024.

\bibitem[Ye et~al.(2025)Ye, Wu, Lu, Chang, Guo, Zhou, Zhao, and Han]{ye2025hi3dgen}
Chongjie Ye, Yushuang Wu, Ziteng Lu, Jiahao Chang, Xiaoyang Guo, Jiaqing Zhou, Hao Zhao, and Xiaoguang Han.
\newblock Hi3dgen: High-fidelity 3d geometry generation from images via normal bridging.
\newblock \emph{arXiv preprint arXiv:2503.22236}, 3, 2025.

\bibitem[Zhang et~al.(2024{\natexlab{a}})Zhang, Xiong, and Xu]{zhang20243d}
Jinzhi Zhang, Feng Xiong, and Mu Xu.
\newblock 3d representation in 512-byte: Variational tokenizer is the key for autoregressive 3d generation.
\newblock \emph{arXiv preprint arXiv:2412.02202}, 2024{\natexlab{a}}.

\bibitem[Zhang et~al.(2024{\natexlab{b}})Zhang, Xiong, and Xu]{zhang2024g3pt}
Jinzhi Zhang, Feng Xiong, and Mu Xu.
\newblock G3pt: Unleash the power of autoregressive modeling in 3d generation via cross-scale querying transformer.
\newblock \emph{arXiv preprint arXiv:2409.06322}, 2024{\natexlab{b}}.

\bibitem[Zhang et~al.(2023)Zhang, Rao, and Agrawala]{zhang2023adding}
Lvmin Zhang, Anyi Rao, and Maneesh Agrawala.
\newblock Adding conditional control to text-to-image diffusion models.
\newblock In \emph{Proceedings of the IEEE/CVF international conference on computer vision}, pages 3836--3847, 2023.

\bibitem[Zhang et~al.(2024{\natexlab{c}})Zhang, Wang, Zhang, Qiu, Pang, Jiang, Yang, Xu, and Yu]{zhang2024clay}
Longwen Zhang, Ziyu Wang, Qixuan Zhang, Qiwei Qiu, Anqi Pang, Haoran Jiang, Wei Yang, Lan Xu, and Jingyi Yu.
\newblock Clay: A controllable large-scale generative model for creating high-quality 3d assets.
\newblock \emph{ACM Transactions on Graphics (TOG)}, 43\penalty0 (4):\penalty0 1--20, 2024{\natexlab{c}}.

\bibitem[Zhao et~al.(2025)Zhao, Lai, Lin, Zhao, Liu, Yang, Feng, Yang, Zhang, Yang, et~al.]{zhao2025hunyuan3d}
Zibo Zhao, Zeqiang Lai, Qingxiang Lin, Yunfei Zhao, Haolin Liu, Shuhui Yang, Yifei Feng, Mingxin Yang, Sheng Zhang, Xianghui Yang, et~al.
\newblock Hunyuan3d 2.0: Scaling diffusion models for high resolution textured 3d assets generation.
\newblock \emph{arXiv preprint arXiv:2501.12202}, 2025.

\end{thebibliography}
}
\end{document}